\documentclass{article}

  \usepackage[dblblindworkshop, final]{neurips_2026}

\usepackage[utf8]{inputenc} 
\usepackage[T1]{fontenc}    
\usepackage{hyperref}       
\usepackage{url}            
\usepackage{booktabs}       
\usepackage{amsmath}       
\usepackage{float} 
\usepackage{amsfonts}       
\usepackage{nicefrac}       
\usepackage{microtype}      
\usepackage{xcolor}         

\title{Recovered or Memorized? Investigating Dynamical Consistency and Faithfulness in World Model Latents}

\author{%
  Jerome Francisco \\
  Southern Alberta Institute of Technology\\
  \texttt{edu.sait.ca} \\
 \And
  Kyle Deng \\
  Carleton University\\
  \texttt{cmail.carleton.ca} \\
   \And
  George Flint\thanks{Senior authors} \\
  University of California, Berkeley\\
  \texttt{berkeley.edu} \\
   \And
  Kevin Zhu$^*$ \\
  Algoverse AI Research\\
  \texttt{algoverseairesearch.org} \\   
}

\usepackage{xurl}
\usepackage{placeins}

\begin{document}

\maketitle

\begin{abstract}
World models develop internal simulators that can learn the structure and dynamics of an environment. It has been demonstrated that their internal representations are both meaningful and causal, but are these representations faithful to the governing dynamics of the environment they are learning? We introduce Anchored Recovery Certification (ARC), a procedure that utilizes a discrete Fr\'echet distance calculation with ground truth to evaluate dynamical consistency on candidate world models up to an affine gauge. We use the procedure on TD-MPC2 (decoder-free world model) that has learned the pendulum swing on an observation lift under varying hyperparameters at $k=5,10,25$ steps. The model is then steered within a PCA-derived intervention subspace and compared against a ground-truth simulator using relative error to investigate dynamical faithfulness. ARC sufficiency held for TD-MPC2 most consistently at $k=10$, with six of seven constraints statistically significant after Holm-Bonferroni correction, while the relative error between the simulator and the causal latent remained substantial. Further research could be conducted with decoder world models such as DreamerV3 (reconstruction loss) to investigate whether the results differ. 
\textbf{Code:} \url{https://github.com/jfran03/dynamical-investigation}
\end{abstract}

\section{Introduction}

Recent work suggests that world models are capable of developing linearly decodable representations of their training environments and are causal. Zhang \cite{zhang} was able to demonstrate this using linear and nonlinear probing, causal interventions, and attention analysis on IRIS (discrete token transformer) and DIAMOND (continuous diffusion UNet), trained on Atari Breakout and Pong. Although causal representations are essential for understanding how the world model behaves, how do we know if these internal representations are not just memorizing trajectories, but rather learning the governing dynamics of its environment to produce its expected behavior?

World models support prediction, planning, and control in learned environments \citep{wmsurvey, worldmodel}. We evaluate whether their representations support trajectory-consistent predictions and intervention responses in a controlled pendulum swing-up environment.


This paper proposes Anchored Recovery Certification (ARC), a procedure that uses a discrete Fréchet distance calculation \citep{frechet} with a ground truth to determine if a world model exhibits evidence of dynamical consistency that cannot be explained by a memorizer. The procedure can be intuitively described as follows: A ruler is developed by creating the ceiling and the floor scored by the same Fréchet calculation to measure a candidate world model. The ceiling is defined to be a model that is an exact replication of the ground truth, whereas the floor is described to be a model that behaves poorly based on the ground truth. To give the ruler markings, perturbations are placed along the ruler, providing a reference for candidate models. Using this ruler, we can measure the dynamical performance of different candidate models, namely a trained world model and a memorizer. If a trained world model were to outperform the memorizer under the same conditions, we can further investigate the trained world models by steering their causal latent to check if it contains the same governing dynamics the true environment has.

We used this procedure using TD-MPC2 \citep{tdmpc2}, trained in a pendulum swing environment with an observation lift. Once the model has finished training, an affine gauge is applied during the Fréchet calculation. DMControl \citep{dmcontrol} is used as the simulator to define the pendulum swing's ground truth.

\section{Method}
\label{gen_inst}

\subsection{Observation Lift \& Affine Gauge}

The observation lift is employed to ensure that the procedure can properly assess the dynamical consistency of a trained world model. It is a randomly initialized, frozen MLP with architecture $3 \rightarrow 64 \rightarrow 32$ and $\tanh$ activations, sampled once and used for all training, calibration, and evaluation data. This gives a memorizer enough structure to be used in the procedure without affecting the score of other models. \citep{dysco}

We use an affine gauge $g(z)=Wz+b$ to map latent state $z$ into the lifted observation space. For each seed, $(W,b)$ are fitted by least squares on a held out calibration set $\mathcal{A}$, disjoint from training data and evaluation rollouts.


\subsection{Constructing Anchored Recovery Certification (ARC)}
ARC is constructed using three different types of measurements which need to undergo the same calculations as the candidate world model: \textit{The ceiling, floor, and perturbation ladder}.

The discrete Fréchet distance calculation can be intuitively defined as a man walking a dog on a leash, both are traveling two different curves at two different speeds, we calculate the shortest length of the leash such that both the man and dog are still traversing through their paths without backtracking. The similarity between two curves is calculated and will not punish trained models that may have dynamical consistency to its ground truth, but potentially at a different speed and time. \citep{frechet}

We convert discrete Fr\'echet distance into a Fr\'echet Skill Score (FSS) \citep{skillscore}. For each scoring window $w\in\mathcal{W}$, let $\hat{\tau}^{(w)}_{\mathrm{cand}}$ denote the candidate trajectory and let $\tau^{(w)}_{\mathrm{true}}$ denote the corresponding ground-truth trajectory. We define FSS as

\begin{equation}
\mathrm{FSS}_{\mathrm{cand}}
=
1-
\frac{
\sum_{w\in\mathcal{W}}
d_{\mathrm{F}}\!\left(
\hat{\tau}^{(w)}_{\mathrm{cand}},
\tau^{(w)}_{\mathrm{true}}
\right)
}{
\sum_{w\in\mathcal{W}}
\max_{t\in\{0,\ldots,k\}}
\left\|
\tau^{(w)}_{\mathrm{true},t}
-
\tau^{(w)}_{\mathrm{true},0}
\right\|_2
},
\label{eq:fss}
\end{equation}

where $d_{\mathrm{F}}$ is the discrete Fr\'echet distance. An FSS of $1$ indicates zero Fr\'echet distance from the ground-truth trajectory. An FSS of $0$ indicates that the aggregate trajectory error equals the aggregate ground-truth displacement over the scoring windows, and negative values indicate larger error. We report FSS at horizons $k\in\{5,10,25\}$. Because the denominator in Eq.~1 is itself the discrete Fr\'echet distance between the true trajectory and a trajectory held fixed at the initial state, a null that predicts no motion (Persistence) yields $\mathrm{FSS}=0$ identically, by construction rather than by measurement.



The ceiling is computed through DMControl \citep{dmcontrol}, which is rolled forward $k$ steps using the true dynamics. The simulator will contain the observation lift and its own affine gauge to ensure a controlled measurement is taken. Because the true operators are being scored, this system will provide the highest possible FSS.

All nulls use the trained model's initial states, action sequences, rollout horizons, and scoring windows. The replay buffer is partitioned before evaluation into 25\% alignment data and 75\% probe data. The Memorizer retrieves the Euclidean nearest neighbor in the lifted observation space from the alignment split and iterates its retrieved output. The Inverse Memorizer retrieves the Euclidean nearest neighbor in autoencoder latent space and iterates its decoded output. The Untrained World Model uses the trained-model architecture with random, untrained weights. Persistence repeats the initial observation. Null selection is performed on the alignment split, selecting the null with the highest FSS score; final null and model scores are computed only on the probe split. The floor for a given constraint and horizon is therefore defined as the maximum FSS attained among these four nulls; because the FSS of Persistence is fixed at $0$ by the normalization in Eq.~\ref{eq:fss} rather than estimated from data, a floor equal to $0$ indicates that the three estimated nulls (Memorizer, Inverse Memorizer, Untrained World Model) all fell below this fixed value, not that Persistence was the strongest competitor.


The perturbation ladder is developed by perturbing each term on the angular acceleration formula (Eq.~\ref{eq:pendulum}):

\begin{equation}
    \ddot{\theta} =
    \underbrace{-\omega^{2}\sin\theta}_{\text{gravity}}
    \;\underbrace{-\,\gamma\dot{\theta}}_{\text{damping}}
    \;+\;\underbrace{\kappa u}_{\text{control}}
    \label{eq:pendulum}
\end{equation}

For the perturbation ladder, we evaluate 87 perturbation draws: one unperturbed reference, 54 coefficient perturbations ($3\omega \times 2\gamma \times 9\kappa$), and 32 random direction ladder draws ($4\epsilon$ levels $\times 8$ directions). Reported scores are averaged across these draws. We use $\eta=(2,1,1)$ so that perturbations to gravity have an effect comparable in scale to those of damping and control, which are weighted equally.


\begin{equation}
c_i^{\mathrm{pert}} = c_i^{\mathrm{true}} \cdot \exp(\eta_i \cdot \varepsilon \cdot u_i), \qquad \eta=(2,1,1), \|\mathbf{u}\|_2 = 1
\label{eq:perturbation}
\end{equation}

\subsection{World Model}
TD-MPC2 \citep{tdmpc2} is trained on the pendulum swing environment using the lifted observation state. The model is swept on three constraining hyperparameters to determine whether dynamical consistency would emerge only in certain conditions.

\begin{table}[H]
  \caption{Constraint Hyperparameters of TD-MPC2}
  \label{tab:constraints}
  \centering
  \begin{tabular}{ll}
    \toprule
    Constraint     & (low / mid / high )     \\
    \midrule
    Horizon (rollout length) & 1 / 3 / 5  \\
    Latent Dim (bottleneck) & 128 / 512 / 1024 \\
    Data Augmentation ($\sigma$) & 0 / 0.05 / 0.15 \\
    \bottomrule
  \end{tabular}
\end{table}

We train 35 models across seven constraint conditions and five random seeds per condition. The baseline uses median constraint values, while the remaining conditions vary one constraint at a time. To align latent states with observations, we fit an affine map $g(z)=Wz+b$ by least squares on a subset $\mathcal{A}$ of off-support states \citep{dysco}. We construct $\mathcal{A}$ in the simulator by pushing angular velocity beyond the first 99th-percentile band reachable under random actions. A state is
retained when its Euclidean distance to its nearest replay-buffer state exceeds the fixed threshold of 1.0; all states in $\mathcal{A}$ are excluded from training and reported evaluation rollouts.


\begin{equation}
W^{*},\, b^{*}
\;=\;
\arg\min_{W,\,b}
\sum_{i \,\in\, \mathcal{A}}
\left\lVert W\, z_{i} + b - y_{i} \right\rVert_{2}^{2}.
\end{equation}

The fitted map is used only to project latent rollouts into the lifted observation space for evaluation.

\subsection{Causal Intervention \& Faithfulness Protocol}

Due to TD-MPC2's SimNorm latent architecture \citep{tdmpc2}, we perform principal component analysis (PCA) on encoder state-Jacobians \citep{activesubspace} evaluated on states in $\mathcal{A}$. For each run, subspace $S$ is the minimum set of principal components explaining at least 99\% of the variance; its dimension and explained variance are recorded per run. For each off-support probe, we patch the component of the encoder latent state projected onto $S$ before the model rollout and evaluate the resulting trajectory at $k\in\{5,10,25\}$.


We test faithfulness using randomly sampled off-support probes projected into $S$; intervention magnitudes are therefore determined by the probes rather than fixed globally. We compare the gauge-projected model response with the simulator response over $k\in\{5,10,25\}$ steps:

\begin{equation}
\mathrm{RelErr}_k =
\frac{\left\lVert \Delta\hat{y}_k-\Delta y_k \right\rVert_2}
{\left\lVert \Delta y_k \right\rVert_2+\delta},
\label{eq:relerr}
\end{equation}
where $\Delta\hat{y}_k$ and $\Delta y_k$ are the model and simulator responses, respectively, and $\delta>0$ is fixed.




\section{Results}

\subsection{ARC Sufficiency} By calculating the Fréchet Skill Score (FSS) of the \textit{ceiling and the floor} of a given constraint, we are able to highlight the maximum and minimum possible scores a model can attain. The FSS for the \textit{ceiling} will always be 1, and table~\ref{tab:fss-nulls} indicates the \textit{floor} FSS for a given constraint. A trained world model under the respective constraint must attain an FSS score between the \textit{ceiling} and the \textit{floor} to be considered ARC-sufficient.

Table~\ref{tab:fss-constraint} compares mean trained-model FSS with the corresponding null floors in Table~\ref{tab:fss-nulls}. At $k=5$, six of seven constraints exceed their selected floor on average, though a paired significance test (Table~\ref{tab:sig-test}) confirms this for only two of seven (Baseline, Narrow Latent Dim) after Holm--Bonferroni correction. All seven constraints exceed their selected floor on average at $k=10$, and six of seven remain significant after correction. At $k=25$, only Wide Latent Dim and Long Horizon exceed their listed floors on average, but neither difference is statistically significant (Table~\ref{tab:sig-test}); at this horizon the Memorizer, Inverse Memorizer, and Untrained World Model all score below the fixed value of Persistence, whose FSS is $0$ by construction rather than by measurement (Eq.~\ref{eq:fss}). The listed floor at $k=25$ is therefore not an empirical measurement, and ARC sufficiency cannot be meaningfully certified against it at this horizon. Thus, ARC sufficiency is strongest and most reliably supported at $k=10$ (six of seven constraints significant), partially supported at $k=5$ (two of seven), and statistically unsupported at $k=25$ (zero of seven) despite two constraints appearing to exceed the floor on raw means. These results indicate that the trained models can produce trajectories that are more consistent with the ground-truth simulator than the selected null models under the conditions in which they significantly exceed the corresponding floor. Because discrete Fréchet distance permits monotone temporal reparameterization, this result should be interpreted as evidence of trajectory consistency rather than direct recovery of the pendulum's governing operators.



\begin{table}[htbp]
  \caption{FSS on TD-MPC2 Constraints, based on a given constraint and $k$ step horizon, with five seeds per constraint. However, the trained world models at $k=25$ steps seems to fail in learning the pendulum swing task overall. Values are means across five seeds; uncertainties denote one standard error of the mean.}
  \label{tab:fss-constraint}
  \centering
  \begin{tabular}{llll}
    \toprule
    Constraint&k=5&k=10&k=25\\
    \midrule
    Baseline & 0.6 $\pm0.0420$ & 0.577 $\pm0.0414$ & -0.0132 $\pm{0.110}$ \\
    Short Horizon & 0.615 $\pm0.0387$ & 0.54 $\pm0.0435$ & -0.421 $\pm{0.0846}$ \\
    Long Horizon & 0.539 $\pm0.0534$ & 0.523 $\pm0.0568$ & 0.0682 $\pm{0.0848}$\\
    Narrow Latent Dim & 0.744 $\pm0.0590$ & 0.642 $\pm0.0237$ & -0.105 $\pm{0.0489}$\\
    Wide Latent Dim & 0.571 $\pm0.0237$ & 0.581 $\pm{0.0224}$ & 0.106 $\pm{0.104}$\\
    No Data Augmentation & 0.431 $\pm0.110$ & 0.403 $\pm{0.141}$ & -0.171 $\pm{0.219}$\\
    High Data Augmentation & 0.445 $\pm0.0750$ & 0.379 $\pm{0.0704}$ & -0.564 $\pm{0.130}$\\
    \bottomrule
  \end{tabular}
\end{table}

\begin{table}[htbp]
\centering
\caption{FSS on TD-MPC2 Constraint Floors, based on a given constraint and $k$ step horizon, with five seeds per constraint. FSS values at $k=\{5,10\}$ are memorizer floors, whereas the FSS values at $k=25$ are persistence floors. Values are means across five seeds; uncertainties denote one standard error of the mean.}
\label{tab:fss-nulls}
\begin{tabular}{lccc}
\toprule
Constraint & $k=5$ & $k=10$ & $k=25$ \\
\midrule
Baseline Null              & $0.465 \pm 0.0288$ & $0.277 \pm 0.0335$ & $0.0000\pm 0.0000^*$   \\
Short Horizon Null         & $0.510 \pm 0.0177$ & $0.297 \pm 0.0365$ & $0.0000 \pm 0.0000^*$   \\
Long Horizon Null          & $0.446 \pm 0.0285$ & $0.253 \pm 0.0388$ & $0.0000 \pm 0.0000^*$ \\
Narrow Latent Dim Null     & $0.477 \pm 0.0276$ & $0.269 \pm 0.0424$ & $0.0000 \pm 0.0000^*$   \\
Wide Latent Dim Null       & $0.520 \pm 0.0239$ & $0.325 \pm 0.0348$ & $0.0000 \pm 0.0000^*$   \\
No Data Augmentation Null  & $0.495 \pm 0.0281$ & $0.269 \pm 0.0584$ & $0.0000 \pm 0.0000^*$   \\
High Data Augmentation Null& $0.325 \pm 0.0284$ & $0.114 \pm 0.0453$ & $0.0000 \pm 0.0000^*$   \\
\bottomrule
\multicolumn{4}{l}{\small * The three estimated nulls scored below 0, this is Persistence and not a measured floor}
\end{tabular}
\end{table}

\begin{table}[htbp]
\centering
\caption{Paired one-sided $t$-test of the mean FSS difference between each TD-MPC2 constraint and its floor, based on a given constraint and $k$ step horizon, with five paired seeds per constraint. Significance reflects a Holm-Bonferroni correction applied across all 21 tests (7 constraints $\times$ 3 horizons). At $k=\{5,10\}$, the floor is the maximum FSS attained among the estimated nulls (Table~\ref{tab:fss-nulls}); at $k=25$, the floor is the persistence floor, which is zero by construction rather than measured. Values are mean paired differences ($\Delta$)  with 95\% confidence intervals.}
\label{tab:sig-test}
\begin{tabular}{lccc}
\toprule
Constraint & $k=5$ & $k=10$ & $k=25^{\dagger}$ \\
\midrule
Baseline & $0.135\,[0.074,\,0.195]^{*}$ & $0.300\,[0.219,\,0.381]^{*}$ & $-0.013\,[-0.320,\,0.293]$ \\
Short Horizon & $0.105\,[0.035,\,0.174]$ & $0.243\,[0.140,\,0.345]^{*}$ & $-0.421\,[-0.656,\,-0.186]$ \\
Long Horizon & $0.093\,[0.022,\,0.164]$ & $0.270\,[0.220,\,0.321]^{*}$ & $0.068\,[-0.167,\,0.304]$ \\
Narrow Latent Dim & $0.267\,[0.129,\,0.404]^{*}$ & $0.372\,[0.268,\,0.477]^{*}$ & $-0.105\,[-0.241,\,0.031]$ \\
Wide Latent Dim & $0.051\,[-0.031,\,0.134]$ & $0.256\,[0.186,\,0.327]^{*}$ & $0.105\,[-0.185,\,0.394]$ \\
No Data Augmentation & $-0.064\,[-0.374,\,0.246]$ & $0.134\,[-0.271,\,0.539]$ & $-0.017\,[-0.624,\,0.590]$ \\
High Data Augmentation & $0.124\,[-0.055,\,0.303]$ & $0.264\,[0.144,\,0.384]^{*}$ & $-0.564\,[-0.927,\,-0.201]$ \\
\bottomrule
\multicolumn{4}{l}{\small * significant after Holm--Bonferroni correction across all 21 tests ($\alpha=0.05$, one-sided).} \\
\multicolumn{4}{l}{\small $^{\dagger}$ $k=25$ tests the model against the fixed value 0 (Persistence floor), not an estimated floor.} \\
\end{tabular}
\end{table}

\begin{table}[htbp]
  \centering
  \caption{FSS for the TD-MPC2 perturbation ladder at $k=10$. The ladder evaluates ground-truth simulators with controlled perturbations to the pendulum-dynamics coefficients and provides a diagnostic reference for interpreting the magnitude of trained-model FSS values. Similar FSS values do not imply recovery of a corresponding fraction of the governing dynamics. Values are means across five seeds; uncertainties denote one standard error of the mean.}
  \label{tab:pert-ladder}
  \begin{tabular}{lcccc}
    \toprule
    Constraint & $\epsilon=0.05$ & $\epsilon=0.1$ & $\epsilon=0.2$ & $\epsilon=0.5$ \\
    \midrule
    Baseline        & $0.945 \pm 0.00400$ & $0.896 \pm 0.00648$ & $0.810 \pm 0.0120$ & $0.589 \pm 0.0551$ \\
    Short Horizon     & $0.945 \pm 0.00388$ & $0.897 \pm 0.00638$ & $0.812 \pm 0.0121$ & $0.595 \pm 0.0541$ \\
    Long Horizon    & $0.944 \pm 0.00468$ & $0.895 \pm 0.00745$ & $0.808 \pm 0.0134$ & $0.586 \pm 0.0551$ \\
    Narrow Latent Dim  & $0.945 \pm 0.00352$ & $0.896 \pm 0.00574$ & $0.810 \pm 0.0122$ & $0.591 \pm 0.0572$ \\
    Wide Latent Dim & $0.944 \pm 0.00392$ & $0.894 \pm 0.00637$ & $0.808 \pm 0.0122$ & $0.590 \pm 0.0541$ \\
    No Data\\Augmentation     & $0.944 \pm 0.00447$ & $0.895 \pm 0.00731$ & $0.809 \pm 0.0135$ & $0.588 \pm 0.0556$ \\
    High Data\\Augmentation     & $0.942 \pm 0.00504$ & $0.891 \pm 0.00816$ & $0.802 \pm 0.0145$ & $0.581 \pm 0.0550$ \\
    \bottomrule
  \end{tabular}
\end{table}

\FloatBarrier
\subsection{Steering Subspace Faithfulness}
Table~\ref{tab:relerr} shows the relative error based on the response of a state nudge between the trained world model (predicted value) and a pendulum with the true governing dynamics (expected value). For reference, model where a nudge would have no effect on its outcome (worse case) would evaluate a relative error of 1.

Results show that more often than not, at all $k$ steps, the average relative error for all constraints of a trained world model is greater than 1, meaning that a predicted response to a state nudge is worse than predicting no response at all. At $k=10$ and $k=25$, the mean relative error exceeds 1 for all evaluated constraints. This indicates that the tested interventions do not reliably reproduce the simulator response at these horizons. The findings are conditional on the affine gauge, PCA-derived subspace, and activation patching protocol. They do not exclude the possibility that relevant dynamics are encoded nonlinearly or outside the selected subspace.

\begin{table}[htbp]
  \centering
  \caption{Average relative error between the gauge-predicted model response and the ground-truth simulator response after interventions within subspace $S$, evaluated on off-support states. Values are means across five seeds.} 
  \label{tab:relerr}
  \begin{tabular}{lccc}  \toprule
  Constraint & $k=5$ & $k=10$ & $k=25$ \\
  \midrule
  Baseline               & 1.00 & 1.61 & 2.09 \\
  Short Horizon           & 0.94 & 1.50 & 2.27 \\
  Long Horizon            & 1.05 & 1.57 & 1.95 \\
  Narrow Latent Dim       & 0.92 & 1.38 & 1.93 \\
  Wide Latent Dim         & 0.89 & 1.42 & 1.89 \\
  No Data Augmentation    & 0.84 & 1.29 & 1.73 \\
  High Data Augmentation  & 1.03 & 1.45 & 2.03 \\
  \bottomrule
  \end{tabular}
  \end{table}

\FloatBarrier
\section{Discussion and Conclusion}
Our results show that TD-MPC2 exhibits trajectory consistency with the pendulum simulator under some constraints and horizons, most consistently at \(k=10\). However, the intervention results do not establish that its internal representation faithfully recovers the pendulum's governing operators.

The decoder-free architecture of TD-MPC2 \citep{tdmpc2} may be the cause of the results. The model would predict its next latent against its encoder's embedding of the true next observation as opposed to reconstruction loss.

\paragraph{Limitations} Our paper has not investigated whether a similar result would occur with world models containing decoders such as DreamerV3, \citep{dreamerv3} which are trained using reconstruction loss. A comparison of two different world models would strengthen the investigation.

Another limitation comes from the affine gauge used. If TD-MPC2 learned the governing dynamics of the pendulum swing nonlinearly, we may have observed that TD-MPC2 contains lower FSS scores than intended. This was an accepted limitation, mitigated as much as possible, to properly convert TD-MPC2's latent into a predicted observation.

\paragraph{Societal Impacts} This approach can advance the scientific understanding of the internal representations in world models, enabling transparency, alignment, and bias detection with respect to the dynamical consistency and faithfulness of a world model. Insight into a model's latent, however, can also enable malicious manipulation, misinformation, or falsified confidence in tools relying on world models. Advisory from ethicists, social scientists, and policy experts are recommended suggestions to take.

\bibliographystyle{plain}
\bibliography{sources.bib}




\appendix

\section{Detailed Per-Seed Results and Null Baselines}

\begin{table}[h]
\centering
\caption{Fr\'echet Skill Score on TD-MPC2 Constraints $k=5$. Values are means across five seeds; uncertainties denote one standard error of the mean.}
\label{tab:fss-constraints-k5}
\begin{tabular}{lcccccc}
\toprule
Constraint & Seed 1 & Seed 2 & Seed 3 & Seed 4 & Seed 5 & Uncertainty \\
\midrule
Baseline               & 0.499 & 0.727 & 0.561 & 0.546 & 0.665 & $\pm 0.0418$ \\
Short Horizon           & 0.632 & 0.532 & 0.715 & 0.519 & 0.675 & $\pm 0.0387$ \\
Long Horizon            & 0.41  & 0.447 & 0.514 & 0.681 & 0.645 & $\pm 0.0534$ \\
Narrow Latent Dim       & 0.671 & 0.738 & 0.962 & 0.615 & 0.733 & $\pm 0.0590$ \\
Wide Latent Dim         & 0.581 & 0.626 & 0.615 & 0.528 & 0.505 & $\pm 0.0237$ \\
No Data Augmentation    & 0.027 & 0.591 & 0.46  & 0.421 & 0.657 & $\pm 0.1097$ \\
High Data Augmentation  & 0.456 & 0.587 & 0.339 & 0.232 & 0.634 & $\pm 0.0750$ \\
\bottomrule
\end{tabular}
\end{table}

\begin{table}[H]
  \caption{Fréchet Skill Score on TD-MPC2 Constraints $k$=10. Values are means across five seeds; uncertainties denote one standard error of the mean.}
  \label{tab:fss-constraints-k10}
  \centering
  \begin{tabular}{lllllll}
    \toprule
    Constraint&Seed 1&Seed 2&Seed 3&Seed 4&Seed 5&Uncertainty\\
    \midrule
    Baseline & 0.492 & 0.684 & 0.522 & 0.517 & 0.671& $\pm0.0414$ \\
    Short Horizon & 0.541 & 0.521 & 0.64 & 0.389 & 0.609& $\pm0.0435$ \\
    Long Horizon & 0.386 & 0.427 & 0.495 & 0.681 & 0.627 &$\pm0.0568$ \\
    Narrow Latent Dim & 0.608 & 0.67 & 0.658 & 0.568 & 0.702& $\pm0.0237$ \\
    Wide Latent Dim & 0.573 & 0.592 & 0.658 & 0.521 & 0.562&$\pm0.0224$ \\
    No Data Augmentation & -0.105 & 0.577 & 0.475 & 0.349 & 0.717&$\pm0.141$ \\
    High Data Augmentation & 0.336 & 0.509 & 0.254 & 0.218 & 0.576 &$\pm0.0704$\\
    \bottomrule
  \end{tabular}
\end{table}

\begin{table}[h]
\centering
\caption{Fr\'echet Skill Score on TD-MPC2 Constraints $k=25$. Values are means across five seeds; uncertainties denote one standard error of the mean.}
\label{tab:fss-constraints-k25}
\begin{tabular}{lcccccc}
\toprule
Constraint & Seed 1 & Seed 2 & Seed 3 & Seed 4 & Seed 5 & Uncertainty \\
\midrule
Baseline               & 0.145   & -0.362 & -0.003  & -0.124 & 0.278  & $\pm 0.1104$ \\
Short Horizon           & -0.581  & -0.32  & -0.388  & -0.638 & -0.178 & $\pm 0.0846$ \\
Long Horizon            & -0.0746 & -0.162 & 0.179   & 0.308  & 0.0907 & $\pm 0.0848$ \\
Narrow Latent Dim       & -0.0566 & -0.0701& -0.00652& -0.291 & -0.102 & $\pm 0.0489$ \\
Wide Latent Dim         & 0.145   & 0.0999 & 0.195   & -0.274 & 0.357  & $\pm 0.1041$ \\
No Data Augmentation    & -0.671  & 0.287  & -0.0225 & -0.271 & 0.592  & $\pm 0.2186$ \\
High Data Augmentation  & -0.763  & -0.304 & -0.755  & -0.806 & -0.191 & $\pm 0.1306$ \\
\bottomrule
\end{tabular}
\end{table}

\begin{table}[h]
\centering
\caption{Fr\'echet Skill Score on TD-MPC2 Constraint Floors $k=5$. Values are means across five seeds; uncertainties denote one standard error of the mean.}
\label{tab:fss-constraint-nulls-k5}
\begin{tabular}{lcccccc}
\toprule
Constraint & Seed 1 & Seed 2 & Seed 3 & Seed 4 & Seed 5 & Uncertainty \\
\midrule
Baseline Null              & 0.412 & 0.531 & 0.393 & 0.459 & 0.53  & $\pm 0.0288$ \\
Short Horizon Null          & 0.491 & 0.464 & 0.563 & 0.495 & 0.537 & $\pm 0.0177$ \\
Long Horizon Null           & 0.389 & 0.394 & 0.418 & 0.524 & 0.505 & $\pm 0.0285$ \\
Narrow Latent Dim Null      & 0.382 & 0.469 & 0.527 & 0.471 & 0.537 & $\pm 0.0276$ \\
Wide Latent Dim Null        & 0.442 & 0.523 & 0.593 & 0.515 & 0.525 & $\pm 0.0239$ \\
No Data Augmentation Null   & 0.517 & 0.492 & 0.395 & 0.505 & 0.567 & $\pm 0.0281$ \\
High Data Augmentation Null & 0.245 & 0.352 & 0.289 & 0.328 & 0.413 & $\pm 0.0284$ \\
\bottomrule
\end{tabular}
\end{table}

\begin{table}[H]
  \caption{Fr\'echet Skill Score on TD-MPC2 Constraint Floors $k=10$. Values are means across five seeds; uncertainties denote one standard error of the mean.}
  \label{tab:fss-constraint-nulls-k10}
  \centering
  \begin{tabular}{lllllllll}
    \toprule
    Constraint&Seed 1&Seed 2&Seed 3&Seed 4&Seed 5&Uncertainty\\
    \midrule
    Baseline & 0.204 & 0.289 & 0.206 & 0.303 & 0.383& $\pm0.0335$ \\
    Short Horizon & 0.278 & 0.181 & 0.384 & 0.277 & 0.367& $\pm0.0365$ \\
    Long Horizon & 0.164 & 0.185 & 0.229 & 0.362 & 0.324&$\pm0.0388$ \\
    Narrow Latent Dim & 0.106 & 0.277 & 0.323 & 0.293 & 0.345&$\pm0.0424$ \\
    Wide Latent Dim & 0.226 & 0.322 & 0.431 & 0.283 & 0.363&$\pm0.0348$ \\
    No Data Augmentation & 0.288 & 0.292 & 0.0507 & 0.31 & 0.403&$\pm0.0584$ \\
    High Data Augmentation &-2.22E-16* & 0.153 & 0.0861 & 0.0658 & 0.267&$\pm0.0453$ \\
    \bottomrule
    \multicolumn{4}{l}{\small * Seed 1 for High Data Augmentation is a persistence floor} \\
  \end{tabular}
\end{table}

\begin{table}[h]
\centering
\caption{Fr\'echet Skill Score on TD-MPC2 Constraint Floors $k=25$. Values are means across five seeds; uncertainties denote one standard error of the mean.}
\label{tab:fss-constraint-nulls-k25}
\begin{tabular}{lcccccc}
\toprule
Constraint & Seed 1 & Seed 2 & Seed 3 & Seed 4 & Seed 5 & Uncertainty \\
\midrule
Baseline Null              & 0 & 0 & 0 & 0      & 0 & $\pm 0.0000$  \\
Short Horizon Null          & 0 & 0 & 0 & 0      & 0 & $\pm 0.0000$  \\
Long Horizon Null           & 0 & 0 & 0 & 0 & 0 & $\pm 0.0000$ \\
Narrow Latent Dim Null      & 0 & 0 & 0 & 0      & 0 & $\pm 0.0000$  \\
Wide Latent Dim Null        & 0 & 0 & 0 & 0      & 0 & $\pm 0.0000$  \\
No Data Augmentation Null   & 0 & 0 & 0 & 0      & 0 & $\pm 0.0000$  \\
High Data Augmentation Null & 0 & 0 & 0 & 0      & 0 & $\pm 0.0000$  \\
\bottomrule
\end{tabular}
\end{table}

\begin{table}[H]
  \caption{Inverse Memorizer null on TD-MPC2 constraints, $k=5$. Values are per-seed; uncertainties denote one standard error of the mean across five seeds.}
  \label{tab:fss-constraint-inverse-memorizer-k5}
  \centering
  \begin{tabular}{lllllll}
    \toprule
    Constraint&Seed 1&Seed 2&Seed 3&Seed 4&Seed 5&Uncertainty\\
    \midrule
    Baseline & 0.299 & 0.307 & 0.248 & 0.277 & 0.391 & $\pm0.0240$ \\
    Short Horizon & 0.317 & 0.263 & 0.373 & 0.273 & 0.326 & $\pm0.0199$ \\
    Long Horizon & 0.264 & 0.283 & 0.267 & 0.354 & 0.389 & $\pm0.0254$ \\
    Narrow Latent Dim & 0.261 & 0.308 & 0.315 & 0.29 & 0.364 & $\pm0.0169$ \\
    Wide Latent Dim & 0.323 & 0.339 & 0.34 & 0.218 & 0.391 & $\pm0.0285$ \\
    No Data Augmentation & 0.384 & 0.361 & 0.23 & 0.342 & 0.422 & $\pm0.0323$ \\
    High Data Augmentation & 0.0476 & 0.18 & 0.106 & 0.191 & 0.235 & $\pm0.0334$ \\
    \bottomrule
  \end{tabular}
\end{table}

\begin{table}[H]
  \caption{Inverse Memorizer null on TD-MPC2 constraints, $k=10$. Values are per-seed; uncertainties denote one standard error of the mean across five seeds.}
  \label{tab:fss-constraint-inverse-memorizer-k10}
  \centering
  \begin{tabular}{lllllll}
    \toprule
    Constraint&Seed 1&Seed 2&Seed 3&Seed 4&Seed 5&Uncertainty\\
    \midrule
    Baseline & 0.0727 & -0.0344 & -0.0381 & 0.104 & 0.173 & $\pm0.0408$ \\
    Short Horizon & 0.0708 & -0.0685 & 0.147 & 0.0188 & 0.124 & $\pm0.0387$ \\
    Long Horizon & -0.0159 & 0.0143 & -0.0177 & 0.185 & 0.226 & $\pm0.0526$ \\
    Narrow Latent Dim & -0.0726 & 0.039 & 0.0624 & 0.0849 & 0.152 & $\pm0.0367$ \\
    Wide Latent Dim & 0.0783 & 0.0993 & 0.0998 & -0.0384 & 0.232 & $\pm0.0429$ \\
    No Data Augmentation & 0.113 & 0.129 & -0.196 & 0.134 & 0.207 & $\pm0.0701$ \\
    High Data Augmentation & -0.391 & -0.0907 & -0.226 & -0.154 & 0.00329 & $\pm0.0665$ \\
    \bottomrule
  \end{tabular}
\end{table}

\begin{table}[H]
  \caption{Inverse Memorizer null on TD-MPC2 constraints, $k=25$. Values are per-seed; uncertainties denote one standard error of the mean across five seeds.}
  \label{tab:fss-constraint-inverse-memorizer-k25}
  \centering
  \begin{tabular}{lllllll}
    \toprule
    Constraint&Seed 1&Seed 2&Seed 3&Seed 4&Seed 5&Uncertainty\\
    \midrule
    Baseline & -0.765 & -0.987 & -0.83 & -0.669 & -0.628 & $\pm0.0636$ \\
    Short Horizon & -0.817 & -0.945 & -0.817 & -0.886 & -0.578 & $\pm0.0623$ \\
    Long Horizon & -0.992 & -0.875 & -0.838 & -0.622 & -0.579 & $\pm0.0783$ \\
    Narrow Latent Dim & -1.191 & -0.848 & -0.829 & -0.719 & -0.554 & $\pm0.1046$ \\
    Wide Latent Dim & -0.888 & -0.768 & -0.783 & -0.845 & -0.513 & $\pm0.0654$ \\
    No Data Augmentation & -1.085 & -0.862 & -1.210 & -0.846 & -0.587 & $\pm0.1076$ \\
    High Data Augmentation & -1.269 & -0.853 & -0.992 & -1.236 & -0.682 & $\pm0.1119$ \\
    \bottomrule
  \end{tabular}
\end{table}

\begin{table}[H]
  \caption{Memorizer Max null on TD-MPC2 constraints, $k=5$. Values are per-seed; uncertainties denote one standard error of the mean across five seeds.}
  \label{tab:fss-constraint-memorizer-max-k5}
  \centering
  \begin{tabular}{lllllll}
    \toprule
    Constraint&Seed 1&Seed 2&Seed 3&Seed 4&Seed 5&Uncertainty\\
    \midrule
    Baseline & 0.412 & 0.531 & 0.394 & 0.46 & 0.531 & $\pm0.0288$ \\
    Short Horizon & 0.491 & 0.464 & 0.563 & 0.495 & 0.537 & $\pm0.0177$ \\
    Long Horizon & 0.391 & 0.394 & 0.42 & 0.524 & 0.505 & $\pm0.0282$ \\
    Narrow Latent Dim & 0.382 & 0.469 & 0.527 & 0.471 & 0.537 & $\pm0.0276$ \\
    Wide Latent Dim & 0.442 & 0.523 & 0.593 & 0.516 & 0.525 & $\pm0.0240$ \\
    No Data Augmentation & 0.517 & 0.492 & 0.396 & 0.506 & 0.567 & $\pm0.0280$ \\
    High Data Augmentation & 0.247 & 0.353 & 0.29 & 0.329 & 0.413 & $\pm0.0282$ \\
    \bottomrule
  \end{tabular}
\end{table}

\begin{table}[H]
  \caption{Memorizer Max null on TD-MPC2 constraints, $k=10$. Values are per-seed; uncertainties denote one standard error of the mean across five seeds.}
  \label{tab:fss-constraint-memorizer-max-k10}
  \centering
  \begin{tabular}{lllllll}
    \toprule
    Constraint&Seed 1&Seed 2&Seed 3&Seed 4&Seed 5&Uncertainty\\
    \midrule
    Baseline & 0.204 & 0.29 & 0.207 & 0.306 & 0.383 & $\pm0.0334$ \\
    Short Horizon & 0.279 & 0.181 & 0.384 & 0.277 & 0.367 & $\pm0.0365$ \\
    Long Horizon & 0.168 & 0.186 & 0.232 & 0.362 & 0.326 & $\pm0.0383$ \\
    Narrow Latent Dim & 0.108 & 0.278 & 0.323 & 0.295 & 0.347 & $\pm0.0422$ \\
    Wide Latent Dim & 0.227 & 0.323 & 0.432 & 0.284 & 0.363 & $\pm0.0349$ \\
    No Data Augmentation & 0.289 & 0.293 & 0.0524 & 0.312 & 0.403 & $\pm0.0582$ \\
    High Data Augmentation & -0.0255 & 0.155 & 0.0862 & 0.066 & 0.268 & $\pm0.0489$ \\
    \bottomrule
  \end{tabular}
\end{table}

\begin{table}[H]
  \caption{Memorizer Max null on TD-MPC2 constraints, $k=25$. Values are per-seed; uncertainties denote one standard error of the mean across five seeds.}
  \label{tab:fss-constraint-memorizer-max-k25}
  \centering
  \begin{tabular}{lllllll}
    \toprule
    Constraint&Seed 1&Seed 2&Seed 3&Seed 4&Seed 5&Uncertainty\\
    \midrule
    Baseline & -0.625 & -0.843 & -0.617 & -0.544 & -0.408 & $\pm0.0705$ \\
    Short Horizon & -0.763 & -0.868 & -0.629 & -0.821 & -0.468 & $\pm0.0726$ \\
    Long Horizon & -0.859 & -0.794 & -0.593 & -0.558 & -0.46 & $\pm0.0750$ \\
    Narrow Latent Dim & -1.063 & -0.59 & -0.714 & -0.651 & -0.399 & $\pm0.1085$ \\
    Wide Latent Dim & -0.759 & -0.605 & -0.574 & -0.872 & -0.381 & $\pm0.0838$ \\
    No Data Augmentation & -0.882 & -0.67 & -1.029 & -0.594 & -0.351 & $\pm0.1173$ \\
    High Data Augmentation & -1.067 & -0.733 & -0.792 & -1.076 & -0.513 & $\pm0.1067$ \\
    \bottomrule
  \end{tabular}
\end{table}

\begin{table}[H]
  \caption{Persistence null on TD-MPC2 constraints, $k=5$. Values are per-seed; uncertainties denote one standard error of the mean across five seeds.}
  \label{tab:fss-constraint-persistence-k5}
  \centering
  \begin{tabular}{lllllll}
    \toprule
    Constraint&Seed 1&Seed 2&Seed 3&Seed 4&Seed 5&Uncertainty\\
    \midrule
    Baseline & -2.22E-16 & -8.88E-16 & -4.44E-16 & -2.22E-16 & -4.44E-16 & $\pm0.0000$ \\
    Short Horizon & 2.22E-16 & -4.44E-16 & -4.44E-16 & -6.66E-16 & 0 & $\pm0.0000$ \\
    Long Horizon & -6.66E-16 & -2.22E-16 & -6.66E-16 & -4.44E-16 & -2.22E-16 & $\pm0.0000$ \\
    Narrow Latent Dim & -2.22E-16 & -4.44E-16 & -2.22E-16 & -2.22E-16 & -4.44E-16 & $\pm0.0000$ \\
    Wide Latent Dim & 0 & -2.22E-16 & 0 & 2.22E-16 & -2.22E-16 & $\pm0.0000$ \\
    No Data Augmentation & -2.22E-16 & -2.22E-16 & -4.44E-16 & -4.44E-16 & -2.22E-16 & $\pm0.0000$ \\
    High Data Augmentation & -2.22E-16 & -2.22E-16 & -6.66E-16 & -6.66E-16 & -1.33E-15 & $\pm0.0000$ \\
    \bottomrule
  \end{tabular}
\end{table}

\begin{table}[H]
  \caption{Persistence null on TD-MPC2 constraints, $k=10$. Values are per-seed; uncertainties denote one standard error of the mean across five seeds.}
  \label{tab:fss-constraint-persistence-k10}
  \centering
  \begin{tabular}{lllllll}
    \toprule
    Constraint&Seed 1&Seed 2&Seed 3&Seed 4&Seed 5&Uncertainty\\
    \midrule
    Baseline & 0 & -8.88E-16 & -2.22E-16 & -4.44E-16 & 0 & $\pm0.0000$ \\
    Short Horizon & 2.22E-16 & -6.66E-16 & -4.44E-16 & -4.44E-16 & -2.22E-16 & $\pm0.0000$ \\
    Long Horizon & -6.66E-16 & -4.44E-16 & -4.44E-16 & -4.44E-16 & -2.22E-16 & $\pm0.0000$ \\
    Narrow Latent Dim & -4.44E-16 & -4.44E-16 & 2.22E-16 & -4.44E-16 & -2.22E-16 & $\pm0.0000$ \\
    Wide Latent Dim & 0 & 0 & 0 & 3.33E-16 & -2.22E-16 & $\pm0.0000$ \\
    No Data Augmentation & -4.44E-16 & -4.44E-16 & -4.44E-16 & -4.44E-16 & 3.33E-16 & $\pm0.0000$ \\
    High Data Augmentation & -2.22E-16 & -2.22E-16 & -4.44E-16 & -4.44E-16 & -6.66E-16 & $\pm0.0000$ \\
    \bottomrule
  \end{tabular}
\end{table}

\begin{table}[H]
  \caption{Persistence null on TD-MPC2 constraints, $k=25$. Values are per-seed; uncertainties denote one standard error of the mean across five seeds.}
  \label{tab:fss-constraint-persistence-k25}
  \centering
  \begin{tabular}{lllllll}
    \toprule
    Constraint&Seed 1&Seed 2&Seed 3&Seed 4&Seed 5&Uncertainty\\
    \midrule
    Baseline & 2.22E-16 & -2.22E-16 & 0 & -2.22E-16 & 2.22E-16 & $\pm0.0000$ \\
    Short Horizon & 2.22E-16 & -4.44E-16 & -2.22E-16 & -4.44E-16 & -2.22E-16 & $\pm0.0000$ \\
    Long Horizon & -6.66E-16 & -2.22E-16 & 0 & -4.44E-16 & 0 & $\pm0.0000$ \\
    Narrow Latent Dim & 1.11E-16 & -2.22E-16 & 2.22E-16 & -2.22E-16 & 2.22E-16 & $\pm0.0000$ \\
    Wide Latent Dim & 0 & -2.22E-16 & 1.11E-16 & 5.55E-16 & -4.44E-16 & $\pm0.0000$ \\
    No Data Augmentation & -2.22E-16 & 0 & 0 & -2.22E-16 & 4.44E-16 & $\pm0.0000$ \\
    High Data Augmentation & -2.22E-16 & 0 & 0 & 8.88E-16 & 4.44E-16 & $\pm0.0000$ \\
    \bottomrule
  \end{tabular}
\end{table}

\begin{table}[H]
  \caption{Untrained Max null on TD-MPC2 constraints, $k=5$. Values are per-seed; uncertainties denote one standard error of the mean across five seeds.}
  \label{tab:fss-constraint-untrained-max-k5}
  \centering
  \begin{tabular}{lllllll}
    \toprule
    Constraint&Seed 1&Seed 2&Seed 3&Seed 4&Seed 5&Uncertainty\\
    \midrule
    Baseline & -5.399 & -6.135 & -5.823 & -5.719 & -5.458 & $\pm0.1330$ \\
    Short Horizon & -6.159 & -5.206 & -5.822 & -6.239 & -4.675 & $\pm0.2982$ \\
    Long Horizon & -5.169 & -5.265 & -5.734 & -5.463 & -4.878 & $\pm0.1433$ \\
    Narrow Latent Dim & -4.779 & -4.580 & -4.502 & -4.487 & -3.563 & $\pm0.2112$ \\
    Wide Latent Dim & -2.712 & -2.518 & -3.002 & -2.490 & -3.000 & $\pm0.1115$ \\
    No Data Augmentation & -5.523 & -5.334 & -6.451 & -5.902 & -4.870 & $\pm0.2667$ \\
    High Data Augmentation & -5.282 & -5.582 & -5.954 & -6.230 & -4.770 & $\pm0.2555$ \\
    \bottomrule
  \end{tabular}
\end{table}

\begin{table}[H]
  \caption{Untrained Max null on TD-MPC2 constraints, $k=10$. Values are per-seed; uncertainties denote one standard error of the mean across five seeds.}
  \label{tab:fss-constraint-untrained-max-k10}
  \centering
  \begin{tabular}{lllllll}
    \toprule
    Constraint&Seed 1&Seed 2&Seed 3&Seed 4&Seed 5&Uncertainty\\
    \midrule
    Baseline & -3.449 & -3.985 & -3.738 & -3.829 & -3.343 & $\pm0.1193$ \\
    Short Horizon & -4.039 & -3.365 & -3.791 & -4.087 & -2.805 & $\pm0.2401$ \\
    Long Horizon & -3.438 & -3.341 & -3.632 & -3.503 & -2.932 & $\pm0.1192$ \\
    Narrow Latent Dim & -3.076 & -2.924 & -2.899 & -2.825 & -2.090 & $\pm0.1731$ \\
    Wide Latent Dim & -1.587 & -1.409 & -1.742 & -1.398 & -1.650 & $\pm0.0674$ \\
    No Data Augmentation & -3.541 & -3.375 & -4.187 & -3.814 & -2.868 & $\pm0.2202$ \\
    High Data Augmentation & -3.405 & -3.442 & -3.721 & -4.038 & -2.993 & $\pm0.1741$ \\
    \bottomrule
  \end{tabular}
\end{table}

\begin{table}[H]
  \caption{Untrained Max null on TD-MPC2 constraints, $k=25$. Values are per-seed; uncertainties denote one standard error of the mean across five seeds.}
  \label{tab:fss-constraint-untrained-max-k25}
  \centering
  \begin{tabular}{lllllll}
    \toprule
    Constraint&Seed 1&Seed 2&Seed 3&Seed 4&Seed 5&Uncertainty\\
    \midrule
    Baseline & -2.886 & -3.285 & -3.069 & -3.268 & -2.552 & $\pm0.1362$ \\
    Short Horizon & -3.473 & -2.853 & -3.176 & -3.413 & -2.202 & $\pm0.2325$ \\
    Long Horizon & -3.064 & -2.857 & -2.907 & -2.930 & -2.210 & $\pm0.1499$ \\
    Narrow Latent Dim & -2.636 & -2.382 & -2.449 & -2.369 & -1.483 & $\pm0.2010$ \\
    Wide Latent Dim & -1.260 & -1.020 & -1.308 & -1.072 & -1.098 & $\pm0.0560$ \\
    No Data Augmentation & -2.982 & -2.830 & -3.339 & -3.220 & -2.077 & $\pm0.2218$ \\
    High Data Augmentation & -2.871 & -2.712 & -2.881 & -3.390 & -2.215 & $\pm0.1881$ \\
    \bottomrule
  \end{tabular}
\end{table}


\end{document}